\documentclass{article}
\usepackage{amsmath,epsfig}
\usepackage{multirow}
\usepackage{hyperref}
\usepackage{url}
\usepackage{graphicx}
\usepackage{dsfont}
\usepackage{caption}
\usepackage{subfigure}
\usepackage{float}
\usepackage{algorithm}
\usepackage{algorithmic}
\usepackage[preprint]{spconfa4}

\let\OLDthebibliography\thebibliography
\renewcommand\thebibliography[1]{
  \OLDthebibliography{#1}
  \setlength{\parskip}{0pt}
  \setlength{\itemsep}{0pt plus 0.3ex}
}

\begin{document}\sloppy

\def\x{{\mathbf x}}
\def\L{{\cal L}}

\title{Boosting few-shot classification with view-learnable contrastive learning}
%
\name{Xu Luo$^{1}$$^{\ast}$, Yuxuan Chen$^{1}$$^{\ast}$, Liangjian Wen$^{1}$$^{\dagger}$, Lili Pan$^{1}$, Zenglin Xu$^{2,3}$$^{\dagger}$}
\address{$^{1}$University of Electronic Science and Technology of China, Chengdu, China\\
$^{2}$Harbin Institute of Technology Shenzhen, Shenzhen, China\\
$^{3}$ Pengcheng Lab, Shenzhen, China\\
Frank.Luox@outlook.com, \{Edmondx.chen, wlj6816\}@gmail.com,\\ lilipan@uestc.edu.cn, xuzenglin@hit.edu.cn}

\maketitle
\renewcommand{\thefootnote}{\fnsymbol{footnote}}
\footnotetext[1]{These authors contributed equally to this work.} 
\footnotetext[2]{Corresponding authors.}

\begin{abstract}

The goal of few-shot classification is to classify new categories with few labeled examples within each class. Nowadays, the excellent performance in handling few-shot classification problems is shown by metric-based meta-learning methods. 
However, it is very hard for previous methods to discriminate the fine-grained sub-categories in the embedding space without fine-grained labels.
This may lead to unsatisfactory generalization to fine-grained sub-categories, and thus affects model interpretation.
To tackle this problem, we introduce the contrastive loss into few-shot classification for learning latent fine-grained structure in the embedding space. Furthermore, to overcome the drawbacks of random image transformation used in current contrastive learning in producing noisy and inaccurate image pairs (i.e., views), we develop a learning-to-learn algorithm to automatically generate different views of the same image. Extensive experiments on standard few-shot learning benchmarks demonstrate the superiority of our method.
\end{abstract}
\begin{keywords}
Few-shot learning, contrastive learning, meta-learning
\end{keywords}
\section{Introduction}
\label{sec:intro}
Few-shot learning has been widely studied to recognize unseen classes with limited samples for each novel class~\cite{DBLP:journals/pami/Fei-FeiFP06,DBLP:conf/icml/FinnAL17,DBLP:conf/cvpr/LeeMRS19,DBLP:conf/iclr/TsengLH020}. Recently, metric-based meta-learning methods have attracted extensive attention in image classification due to their superior performance and simplicity~\cite{DBLP:conf/nips/VinyalsBLKW16,DBLP:conf/nips/SnellSZ17,DBLP:conf/cvpr/SungYZXTH18}. For making inference, these methods compare the similarity between the feature embedding of query images and that of a few labeled images of each class. This therefore requires learning a flexible encoder, which can map the data points with similar semantics in the input space to locate closely in the embedding space. Meanwhile, those data with different semantic meanings in the input space should disperse in the embedding space.
Accordingly, a new sample from the novel class can be recognized directly through a simple distance metric within the learned embedding space. Indeed, the  performance of recognizing novel classes in metric-based meta-learning extremely relies on the learned embedding space.

\begin{figure}[t]
    \centering    
    \includegraphics[width=5cm]{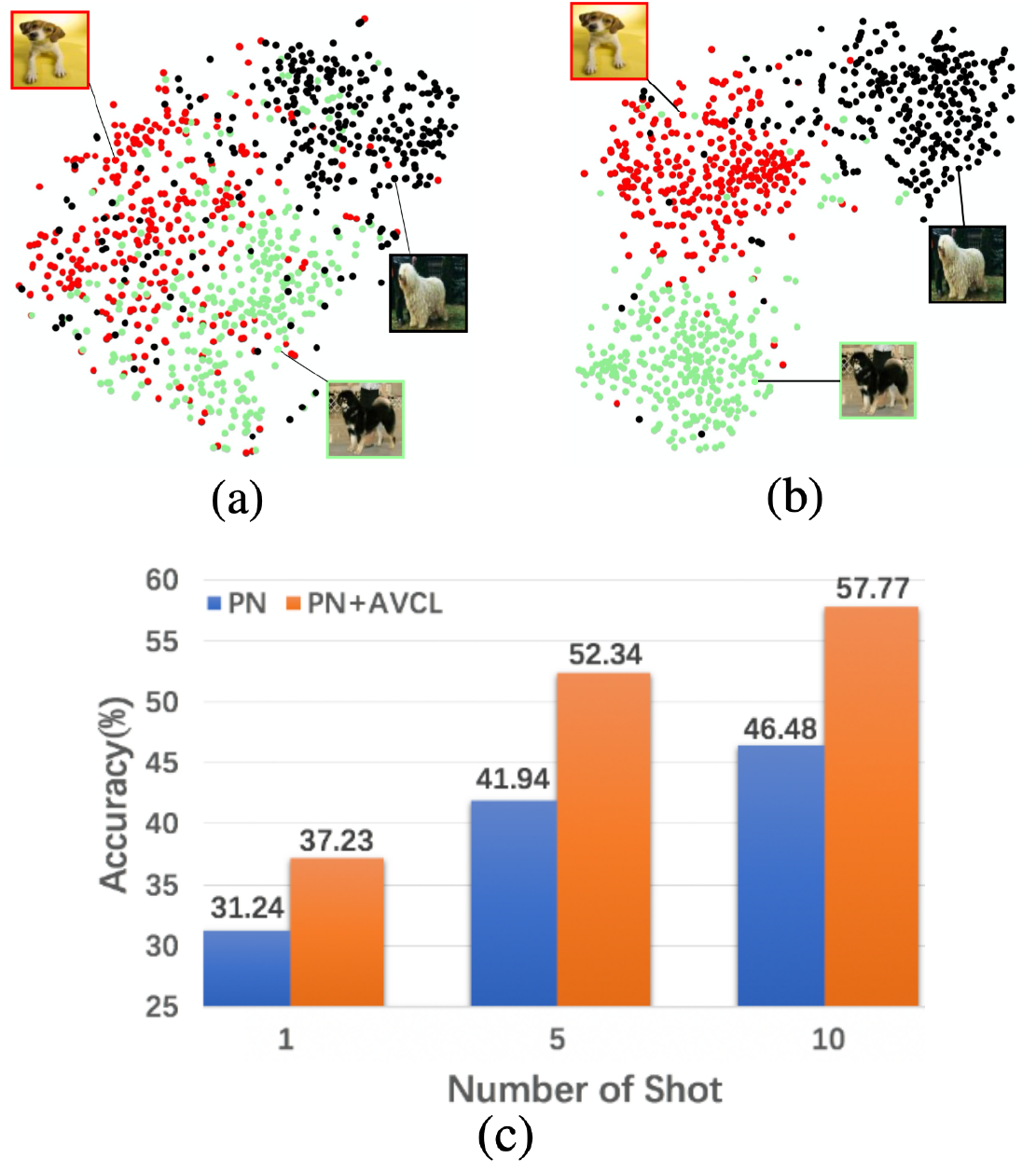}  
 
    \caption{(a) and (b): T-SNE visualization of feature vectors extracted from three out of nine fine-grained classes of dogs in miniImageNet using Prototypical Network and our proposed method, respectively. 
    (c): Evaluation results on the fine-grained  Stanford Dogs dataset~\cite{KhoslaYaoJayadevaprakashFeiFei_FGVC2011}.} 

    \label{intraclass}  
\end{figure}

Despite the success of recognizing novel classes, existing metric-based few-shot approaches often fail to push the fine-grained sub-categories apart in the embedding space given no fine-grained labels in training.
For illustration, we merge nine different sub-categories of dogs in the miniImageNet dataset into a coarse-grained class as a new label ``dog'' to train the Prototypical Network(PN)~\cite{DBLP:conf/nips/SnellSZ17} without changing other classes. As shown in Figure~\ref{intraclass}(a), we visualize the features of the input data of three fine-grained sub-categories of ``dog'' using t-SNE. It is clearly revealed that features of fine-grained classes learned by the Prototypical Network cannot be separated without further label information. This means that these methods often do not generalize well to fine-grained sub-categories. Since labeling the fine-grained sub-categories requires strong expertise, the generalization ability to unseen fine-grained sub-categories is of critical importance.


In this paper, we try to alleviate this problem by seeking self-supervised learning to learn the  fine-grained structure without given corresponding label information. As a powerful self-supervised representation learning paradigm, Contrastive Learning~\cite{DBLP:journals/corr/abs-2002-05709,DBLP:conf/cvpr/He0WXG20}  has outperformed over even supervised learning in many situations. The key insight of contrastive learning is to contrast
semantically similar (positive) with dissimilar (negative) pairs of data points. Nice theoretical results for contrastive learning has been given in \cite{DBLP:conf/icml/SaunshiPAKK19}, by hypothesizing that semantically similar points are often sampled from the same latent class.
 Hence, contrastive learning has the potential to bring closer the representations from the same latent class and to separate those from different latent classes.
Therefore, we introduce a contrastive loss into few-shot classification and learn latent fine-grained structures in the embedding space, which helps to cluster samples with similar representations to form similar sub-categories.

A critical issue in contrastive learning is generating 
 a pair of semantically similar representations(views) of the same image for contrasting with those dissimilar ones  in the embedding space.
However, due to the limited number of training images, random image transformation may generated poor positive pairs with more substantial noises and less concept-relevant information when directly applied to few-shot learning. 
This may therefore make contrastive learning fail to learn fine-grained structure. To effectively improve fine-grained structure learning in the few shot learning setting, we propose view-learnable contrastive learning(VLCL) for metric-based meta-learning. Specifically we replace random image transformation of contrastive learning with spatial transformer network(STN)~\cite{DBLP:conf/nips/JaderbergSZK15}, a learned module that allows flexible spatial manipulation of images, and  develop a learning-to-learn algorithm to adaptively generate different views of the same image. 
In detail, the parameters of STN are optimized through the contribution of the contrastive loss to few-shot image recognition.

To verify that our proposed approach can improve the generalization ability of unseen fine-grained sub-categories without corresponding fine-grained label information, we follow the experiment setting of the coarse-grained class ``dog'' for training and  further test the learned model on Stanford  Dogs, a fine-grain dataset containing 120 fine-grained dog classes~\cite{KhoslaYaoJayadevaprakashFeiFei_FGVC2011}. As shown in Figure~\ref{intraclass}(b,c), our proposed method can learn a better embedding space and significantly improve the test accuracy on the 
Stanford Dogs dataset. For instance, our method obtains an improvement of 10.40\%  on the 5-way 5-shot task over the Prototypical Network.


\section{Related works}
\textbf{Few-shot image recognition:}
Few-shot image recognition was first proposed by~\cite{DBLP:journals/pami/Fei-FeiFP06}, with the aim to solve the problem of classifying novel categories with few labeled images per class. Nowadays, two types of meta-learning methods are the mainstream methods to address this problem. One is gradient-based method that empower the model with ability to rapidly fine-tune to novel classes with limited labeled images ~\cite{DBLP:conf/icml/FinnAL17,DBLP:conf/cvpr/LeeMRS19}. The other is metric-based method, which makes predictions based on a similar metric in a learned feature space between images with and without labels. 
Common similar metric used in this method includes cosine similarity~\cite{DBLP:conf/nips/VinyalsBLKW16}, Euclidean similarity~\cite{DBLP:conf/nips/SnellSZ17}, relation module~\cite{DBLP:conf/cvpr/SungYZXTH18}, and graph neural network~\cite{DBLP:conf/iclr/SatorrasE18}. 

In our work, we primarily consider improving the performance of metric-based meta-learning methods, especially Prototypical Network~\cite{DBLP:conf/nips/SnellSZ17}. Recently, there are many methods proposed to improve the ability of metric-based meta-learning by constraining the structure of the feature space. Li et al.~\cite{DBLP:conf/cvpr/Li0LFLW20} introduce an extra margin loss that leverages external content information, e.g., pre-trained word embeddings, to generate adaptive margin between classes. This leads the feature space to have better semantic structure. Contrastively, our work does not import external information and utilizes semantic information extracted purely from images themselves. Works most relevant to ours is ~\cite{DBLP:conf/iccv/GidarisBKPC19} which utilize self-supervised tasks to improve few-shot learning. However, their self-supervised pretext tasks are fixed at the training stage. Our proposed framework can progressively change the view of contrastive learning under a learning-to-learn paradigm.

\noindent\textbf{Contrastive Learning:}
Contrastive learning is one of the most popular methods for unsupervised visual representation learning. View transformations and contrastive loss are two key parts of contrastive learning. This  framework attains representations by optimizing contrastive loss that maximize agreement between transformed views of the same image and minimize agreement between transformed views of different images. 
Contrastive learning was first proposed by ~\cite{DBLP:conf/cvpr/HadsellCL06}. Recently, Wu et al.~\cite{DBLP:conf/cvpr/WuXYL18} consider instance discrimination and use noise contrastive loss to learn representations. Contrastive multiview Coding~\cite{DBLP:journals/corr/abs-1906-05849} utilizes contrastive learning to attain representations on multi-view setting. SimCLR~\cite{DBLP:journals/corr/abs-2002-05709} summarizes a standard framework of contrastive learning and shows the effect of different random view transformations. However, the performance of SimCLR relies on large batch size. Momentum Contrast (MoCo)~\cite{DBLP:conf/cvpr/He0WXG20} is proposed to alleviate this problem by constructing a queue to preserve immediate preceding samples. In our work, due to the relatively small batch size of metric-based meta-learning, we adopt MoCo in our model. Tian~et al.~\cite{DBLP:journals/corr/abs-2005-10243} also point out the importance of view transformation for contrastive learning, and propose to learn view transformation via information bottleneck principle under unsupervised and semi-supervised settings. Our work employs a learning-to-learn framework to automatically learn views for better performance for few-shot classification.

\section{Proposed Method}

\subsection{Problem Setting}
A few-shot classification task is characterized as N-way, K-shot only if the model adapt to classify new data after seeing K examples from each of the N classes. Meta-learning algorithms imitate this setting in each iteration by randomly sampling N classes and their corresponding images from a base dataset $\mathbf{D}_B$ to construct a pseudo few-shot task or episode $T$. Specifically, the input of each episode $T$ can be divided into two sets: (1)  Observable support set $\mathbf{D}_S=\{(\mathbf{x}_i,\mathbf{y}_i)\}_{i=1}^{N\times K}$, formed by randomly selecting K samples from each of the N classes and (2) Unseen query set $\mathbf{D}_Q=\{(\mathbf{x}_j,\mathbf{y}_j)\}_{j=1}^{M}$, containing other M samples from the same N categories. Given $\mathbf{D}_S$, the pseudo task is employed by making predictions of $\mathbf{D}_Q$. Then the labels of query set are used to compute classification loss to guide the update of the model.   

\subsection{Model description}
\begin{figure*}
     \centering 
     \includegraphics[width=0.6\linewidth]{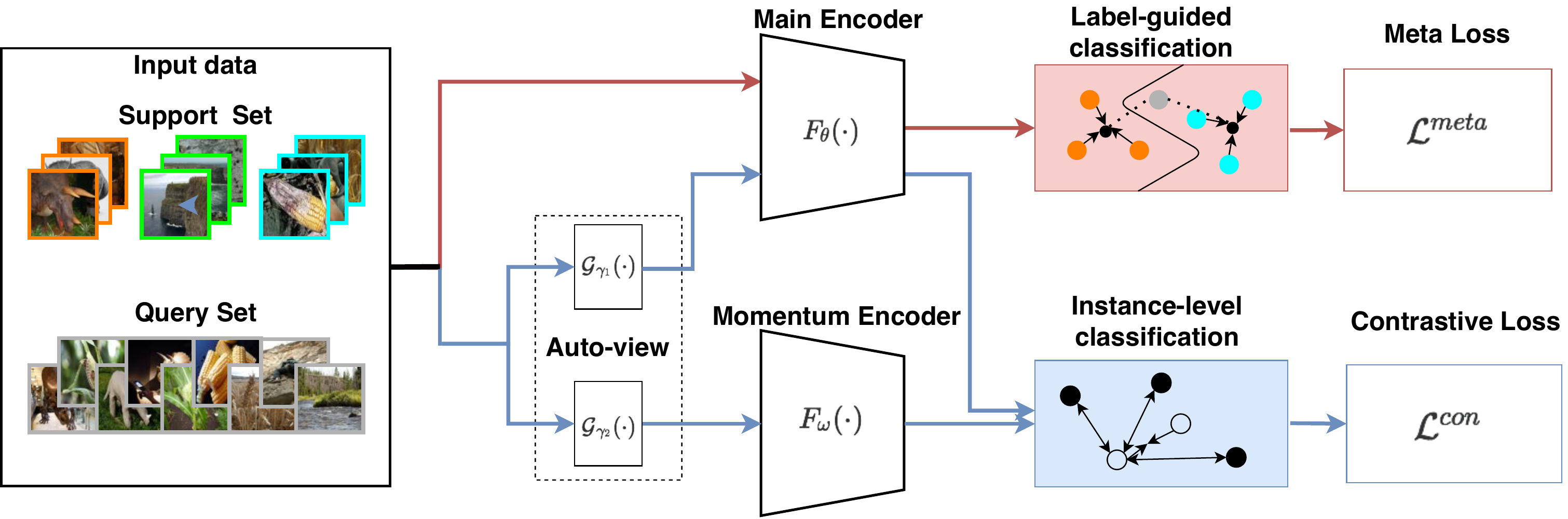}
     \caption{A brief flow diagram of our model at training stage. Our framework consists of two main tasks. The red arrow is metric-based meta-learning, while blue arrow depicts instance-level contrastive learning. At meta-test stage, only the main encoder $F_\theta(\cdot)$ is held for evaluation under few-shot setting.} 
     \label{fig:1}
\end{figure*}

Figure~\ref{fig:1} illustrates the framework of our proposed method. Two complementary classification tasks are employed simultaneously to learn the main encoder $F_\theta(\cdot)$, which is the key component that maps the input into a feature space. One path is metric-based meta-learning, which utilizes explicit label information to regularize the feature space. Another path is contrastive prediction task, which is a self-supervised instance-level classification task. 
This task is designed to  identify latent fine-grained structure in the feature space by aggregating the representations of the same latent class and separating those of different latent classes at the same time.

In an episode T, sampled images in $\mathbf{D}_S$ and $\mathbf{D}_Q$ are used in the two paths. In the metric-based meta-learning path,  the main encoder $F_\theta(\cdot)$ first maps all images into the feature space, and then all support features in the same class $C_i$ aggregate into one vector $h_i$. Typically, this is accomplished by averaging all support features, i.e. $\mathbf{h}_i = \frac{1}{K}\sum_{(x,y)\in \mathbf{D}_S}\mathds{1}_{[y=i]}F_{\theta}(x)$. It is followed by computing the similarities between query features and aggregated features in each class. The final classification loss $\mathcal{L}^{meta}$ is defined as average cross entropy between true labels and predictions based on similarities. This can be formulated as:
\begin{equation}\label{metaloss}
  \mathcal{L}^{\mathrm{meta}}(\mathbf{D}_S,\mathbf{D}_Q,\theta)=-\frac{1}{M}\sum_{(x,y)\in \mathbf{D}_Q}\mathrm{log}\frac{e^{\mathrm{sim}(F_{\theta}(x),h_y))}}{\sum_{i=1}^Ne^{\mathrm{sim}(F_{\theta}(x),h_i)}},
\end{equation}
where $\mathrm{sim}$ denotes a similarity metric. 

At the core of the contrastive path is the process of producing two views from one image which lie in the same latent space, i.e. containing similar semantic contents. This process is accomplished by two differentiable auto-view modules $\mathcal{G}_{\gamma_1}(\cdot)$ and $\mathcal{G}_{\gamma_2}(\cdot)$ parameterized by $\gamma_1$ and $\gamma_2$, respectively. Each of them is a spatial transformer network(STN)~\cite{DBLP:conf/nips/JaderbergSZK15} that allows flexible semantics-invariant spatial manipulation of images, including cropping, one dimensional scaling, As for translation, image deformation and proportional shrinkage, see appendix for more details of STN. They are applied to each image $\mathbf{x}_i$, producing two views $\mathbf{x}_i^{(1)}$ and  $\mathbf{x}_i^{(2)}$. These views are then mapped into feature space by the main encoder $F_{\theta}(\cdot)$ and momentum encoder $F_{\omega}(\cdot)$, respectively. As mentioned in \cite{DBLP:conf/cvpr/He0WXG20}, the momentum encoder's parameter $\omega$ is a moving average of $\theta$, which makes two encoders behave similar, see the appendix for more details of Moco. Given $F_\theta(\mathbf{x}_i^{(1)})$, the contrastive loss aims to identify $F_\omega(\mathbf{x}_i^{(2)})$ in thousands of features $\{F_\omega(\mathbf{x}_k^{(2))}\}_{k\neq i}$, and can be formulated as:
\begin{equation}\label{contrastiveloss1}
\begin{aligned}
    &\mathcal{L}^{\mathrm{con}}(\mathbf{D}_S, \mathbf{D}_Q, \omega, \theta, \gamma) \\
    &= -\sum_{x\in \mathbf{D}_S\cup   \mathbf{D}_Q}\mathrm{log}\frac{e^{\mathrm{sim}(F_\theta(\mathbf{x}^{(1)}),F_\omega(\mathbf{x}^{(2)}))}}{\sum_{j=1}^{r}e^{\mathrm{sim}(F_\theta(\mathbf{x}^{(1)}),F_\omega(\mathbf{x}_j^{(2)})))}},
\end{aligned}
\end{equation}
where $r$ denotes the number of negative samples, and $\gamma$ = [$\gamma_1, \gamma_2$] denotes the parameters of auto-view modules. By
minimizing $\mathcal{L}^{\mathrm{con}}$ w.r.t $\theta$, we force the main encoder $F_{\theta}(\cdot)$ to map  views of one image which are semantically similar into closer points in the feature space, thus constructs a better fine-grained semantic structure.

\subsection{Learning strategy} 
The optimization of our model during each iteration contains two stages. We denote $\theta^t, \omega^t, \gamma^t$ as parameters and $\mathbf{D}_S^t$, $\mathbf{D}_Q^t$ as support set and query set during iteration t, respectively. We first update two encoders based on the meta loss $\mathcal{L}^{\mathrm{meta}}$ and contrastive loss $\mathcal{L}^{\mathrm{con}}$:
\begin{align}\label{encoderupdate}
        \mathcal{L}^{\mathrm{total}}
        &= \mathcal{L}^{\mathrm{meta}}(\mathbf{D}_S^t, \mathbf{D}_Q^t, \theta^{t})+\beta\mathcal{L}^{\mathrm{con}}(\mathbf{D}_S^t, \mathbf{D}_Q^t, \omega^t, \theta^t,\gamma^t), \\
        \theta^{t+1} &=\theta^{t}-\alpha\nabla_{\theta^t}\mathcal{L}^{\mathrm{total}},\\
        \omega^{t+1} &= \epsilon\omega^t+(1-\epsilon)\theta^{t+1},
\end{align}
where $\beta$ denotes the regularization hyperparameter weighting two losses, $\alpha$ denotes the learning rate of $\theta$, and $\epsilon$ denotes momentum coefficient that 
controls the chasing speed of the momentum encoder $F_\omega(\cdot)$. Note that the value of $\theta^{t+1}$ relies on $\gamma^t$ via the contrastive loss $\mathcal{L}^{\mathrm{con}}$. This fact is crucial for the update of $\gamma^t$ in the next stage.

The updating criterion for the auto-view modules $G_{\gamma_1}(\cdot)$ and $G_{\gamma_2}(\cdot)$ is to improve the positive effect of contrastive loss for meta-learning. Keeping this in mind, we use the updated encoder $F_{\theta^{t+1}}(\cdot)$ to compute meta loss again on the same task, and cast the loss as evaluation of update quality in the first stage. Then the loss is used to guide the update of the auto-view modules via gradient descent:
\begin{equation}\label{viewupdate}
    \gamma^{t+1}=\gamma^t-\eta\nabla_{\gamma^t}\mathcal{L}^{\mathrm{meta}}(\mathbf{D}_S^t, \mathbf{D}_Q^t, \theta^{t+1}),
\end{equation}
where $\eta$ is the learning rate of the auto-view modules. We mention again that $\theta^{t+1}$ can be cast as a function of $\gamma^t$. Thus the loss depends on $\gamma^t$ through the computation graph of $\theta^{t+1}$ in first stage. The update of $\gamma^t$ adjusts produced views towards better update of encoder $F_{\theta}(\cdot)$ in the first stage. This indicates that learned views further pushes positive effect of contrastive loss for meta-learning.
 
\section{Experimental Results}
In this section, we conduct experiments to demonstrate the effectiveness of our method. Our proposed method is evaluated on standard few-shot learning benchmarks. We also conduct ablative study about the auto-view module in our learning framework. To further verify the effectiveness of our method, we evaluate our method on three few-shot fine-grained datasets in appendix.


\subsection{Experimental Setup}
We conduct 5-way 5-shot and 5-way 1-shot classification for all datasets. The metric-based meta-learning method adopted in our model is Prototypical Network, one of the state-of-the-art metric-based meta-learning methods for few-shot learning.

\textbf{Datasets:}
We follow the general few-shot image recognition settings and evaluate our method on two benchmarks: \textbf{miniImageNet}~\cite{DBLP:conf/nips/VinyalsBLKW16} and \textbf{CUB-200-2011}~\cite{welinder2010caltech}. miniImageNet is selected from the well-known ImageNet\cite{DBLP:journals/ijcv/RussakovskyDSKS15} dataset. CUB-200-2011 was originally proposed for fine-grained bird classification.  


    


\textbf{Auto-view transformation architectures:}
We use 4-layer convolutional nets with 64 channels as localisation networks in STN modules that receive images of size 80 $\times$ 80 and output 4-dimensional vectors. The vectors are used as parameters of two diagonal affine transformations which apply to images and produce two views. When implementing without auto-view module, we replace the module with a random cropping function followed by resize to original size. Regardless of having this module or not, we randomly apply color jitter, gaussian blur and horizontal flip before it, and the hyperparameters are the same as in~\cite{DBLP:conf/cvpr/He0WXG20}.

\textbf{Feature extractor architectures:}
In all our experiments, the main encoder and momentum encoder share the same architecture. Following prior work~\cite{DBLP:conf/cvpr/GidarisK19,DBLP:conf/iccv/GidarisBKPC19}, we use a 2-layer Wide Residual Network(WRN-28-10) that outputs 640-dimensional feature vectors after global pooling given images(or views) of size 80 $\times$ 80. This feature space is directly used for metric-based meta-learning, but will be further mapped by a 2-layer mlp project head to a 128-dimentional hidden space for contrastive learning.


\textbf{Training details:}
At training, each minibatch contains 4 tasks, and classes for each task are randomly selected from training set. The query set contains 4 samples during meta-training and 16 samples during meta-testing. All samples from query set and support set are used for computing contrastive loss. All learnable components of our model are trained for 60 epochs by SGD optimizer with Nesterov momentum 0.9 and weight decay 0.0005. The learning rate for STN was set to 0.00001, the same magnitude as in the original paper. The learning rate for other parts of model  was initially set to 0.1, and then changed to 0.01 and 0.001 at epochs 20 and 40, respectively. Moreover, regularization hyperparameter $\beta$ was set to 2.0. We use a queue containing 63000 negative samples for contrastive learning. Momentum coefficient $\epsilon$ for updating momentum encoder was set to 0.999, following~\cite{DBLP:conf/cvpr/He0WXG20}.

\subsection{Evaluation on Benchmarks}


\begin{table}[t]

\caption{Comparative results for 5-way  classification on miniImageNet. Average accuracies on the meta-test set with 95 confidence interval are reported. $\dag$ denotes methods using external text information. $\ddag$ denotes result reported in~\cite{DBLP:conf/iccv/GidarisBKPC19}.}
    \label{MiniImageNet}
    \begin{center}
    \begin{tabular}{cc||c|c}
    Model & 1-shot & 5-shot
    \\  \hline
    MAML ~\cite{DBLP:conf/icml/FinnAL17}&  48.70 $\pm$ 1.84 & 63.11 $\pm$ 0.92\\
    Matching Network ~\cite{DBLP:conf/nips/VinyalsBLKW16}   & 43.56 $\pm$ 0.84 &55.31 $\pm$ 0.73 \\
    Relation Networks ~\cite{DBLP:conf/cvpr/SungYZXTH18}   & 50.44 $\pm$ 0.82 & 65.32 $\pm$ 0.70\\
    IDeMe-Net~\cite{DBLP:conf/cvpr/ChenFW00H19}  &  59.14 $\pm$ 0.86 & 74.63 $\pm$ 0.74\\
    

    PPA ~\cite{DBLP:conf/cvpr/QiaoLSY18}    & 59.60 $\pm$ 0.41 & 73.74 $\pm$ 0.19\\
    
     \hline

    PN$^\ddag$ ~\cite{DBLP:conf/nips/SnellSZ17}   & 55.85 $\pm$ 0.48 & 68.72 $\pm$ 0.36\\
    PN+TRAML$^\dag$ ~\cite{DBLP:conf/cvpr/Li0LFLW20}   & 60.31 $\pm$ 0.48 & \textbf{77.94 $\pm$ 0.57}\\
    SEN PN~\cite{DBLP:conf/eccv/kampffmeyersen} &  - & 72.3\\
    PN+rotation ~\cite{DBLP:conf/iccv/GidarisBKPC19}   & 58.28 $\pm$ 0.49 & 72.13 $\pm$ 0.38\\
    \textbf{PN + CL(ours)}   & 59.54 $\pm$ 0.47  & 74.46 $\pm$ 0.52\\
    \textbf{PN + VLCL(ours)}   & \textbf{61.75 $\pm$ 0.43}  & \textbf{76.32 $\pm$ 0.49}
    \\\hline 
    \end{tabular}
    \end{center}
\end{table}

\begin{table}[t]

\caption{Comparative results for 5-way classification on CUB. Average accuracies on the meta-test set with 95 confidence interval are reported.}
    \label{CUB}
    \begin{center}
    \begin{tabular}{c|c||c|c}
    Model  & 1-shot & 5-shot
    \\  \hline
    AFHN \cite{DBLP:conf/cvpr/0012ZL020}  & 70.53 $\pm$ 1.01 & 83.95 $\pm$ 0.63\\
    FEAT \cite{DBLP:journals/corr/abs-1812-03664}  & 68.87 $\pm$ 0.22 & 82.90 $\pm$ 0.15 \\
    MAML \cite{DBLP:conf/iclr/ChenLKWH19}  & 67.28 $\pm$ 1.08 &  83.47 $\pm$ 0.59 \\
    cosine classifier \cite{DBLP:conf/iclr/ChenLKWH19}  & 68.00 $\pm$ 0.83 & 84.50 $\pm$ 0.51 \\
    Relationnet \cite{DBLP:conf/iclr/ChenLKWH19}  & 66.20 $\pm$ 0.99 & 82.30 $\pm$ 0.58 \\
    DEML \cite{DBLP:journals/corr/abs-1812-03664}  & 66.95 $\pm$ 1.06 & 77.11 $\pm$ 0.78\\
    PN \cite{DBLP:conf/nips/SnellSZ17}  & 66.08 $\pm$ 0.54 & 78.79 $\pm$ 0.23 \\
    \textbf{PN + CL(ours)}   & 70.45 $\pm$ 0.41  & 82.67 $\pm$ 0.46\\
    \textbf{PN+VLCL(ours)}   &  \textbf{71.21 $\pm$ 0.43} &
    \textbf{85.08 $\pm$ 0.36} \\ \hline
    \end{tabular}
    \end{center}
\end{table}

    
    

Below we report comparative results on two benchmarks for FSL: MiniImageNet and CUB in Table \ref{MiniImageNet} and \ref{CUB}. In Table \ref{MiniImageNet} we divide methods into two groups and compare them with our proposed method, respectively. The first group contains recent comparative few-shot learning methods. The second group contains baseline method(Prototypical Network) and other methods that aim at improving it. PN+CL denotes models that replace auto-view modules with random cropping functions. Analysis from the results, we can find that: (1)Our method consistently improves the baseline method (prototypical Network). For instance, our model boosts performance of Prototypical Network on miniImageNet by 5.90\% and 7.60\% under the 1-shot and 5-shot settings, respectively. This verifies that our method can indeed improve model performance by refining  fine-grained semantic structure of the feature space . (2)Our model outperforms recent comparable few-shot learning methods and also outperforms other approaches that aim at improving Prototypical Network. Moreover, we achieve competitive performance with the method using external text information (PN+TRAML). This further gives evidence of the superiority of our learned feature space for FSL. (3)Our auto-view module can indeed improve the quality of views, thus reach a better performance. Compared to random views, our VLCL method obtains 2.21\% and 1.86\% performance gains under the 5-way 1-shot and 5-shot settings on miniIMageNet, respectively.

\subsection{Evaluation of view-learnable learning}
\begin{figure}[t]
     \centering 
     \includegraphics[width=1\linewidth]{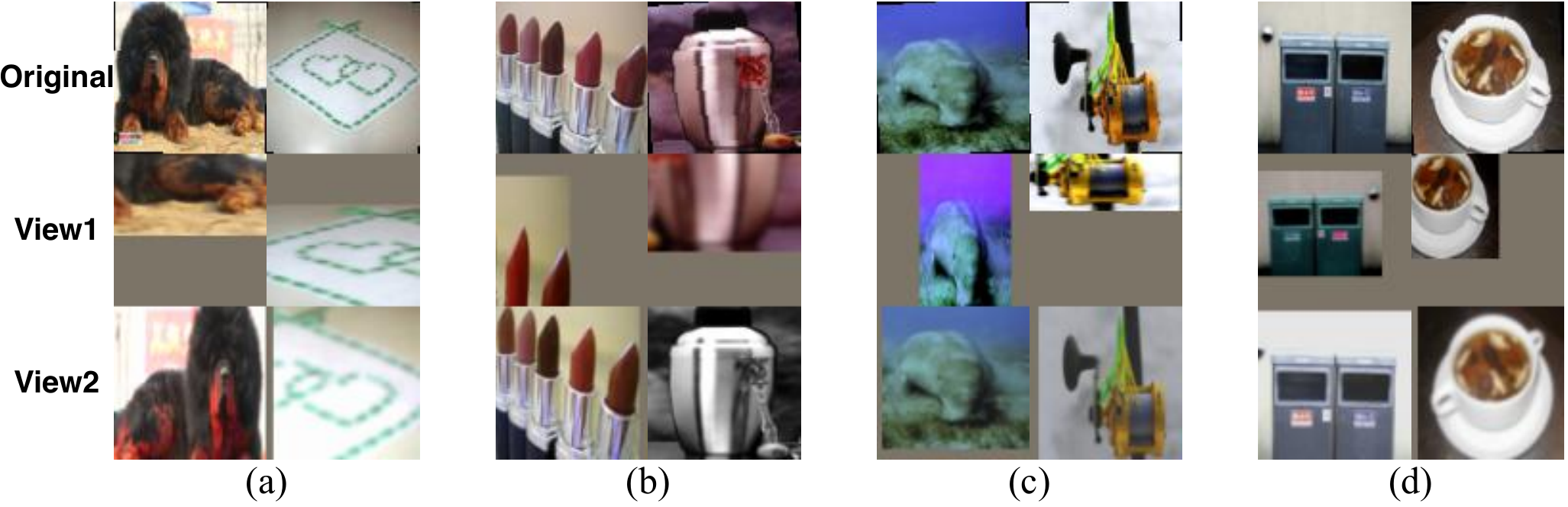}
     \caption{Four types of views learned by our method: (a) Local-to-local, (b) Global-to-local, (c) One-dimension Scaling, and (d) Proportional zooming. } 
     \label{view}
\end{figure}

In Fig.~\ref{view} we show four distinct types of views produced by our auto-view module. We first notice that every pair of learned views are different but both captures the main information of the image, which is expected~\cite{DBLP:journals/corr/abs-2005-10243}. We further notice that, aside from local-to-local and global-to-local views which can also be accomplished by traditional random cropping, our auto-view module additionally allows one dimension scaling, 
translation transformation, image deformation and proportional shrinkage. This flexibility can enrich semantics-invariant transformations applied to the images, forcing the encoder to extract essential content of the image. This allows the samples in the feature space to be distributed according to their semantics. Thus images from the same novel category can be mapped to close points in the feature space, which greatly improves generalization capability. 

\begin{figure}[t]
     \centering 
     \includegraphics[width=0.8\linewidth]{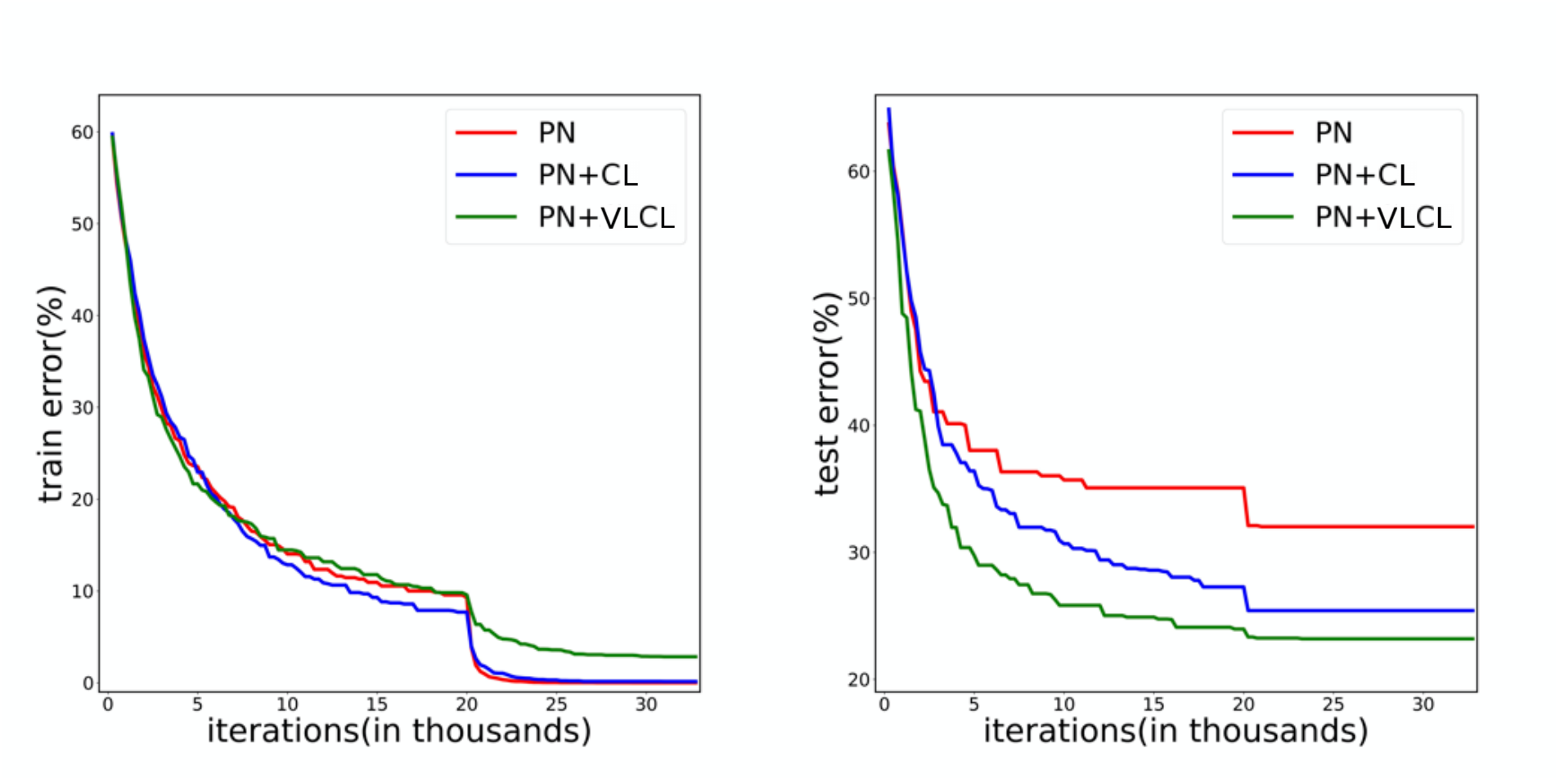}
     \caption{Training errors (left) and test errors (right) on miniImageNet. The auto-view module significantly decreases test errors, while keeping not overfit to the training set.}
     \label{loss}
\end{figure}

 We additionally show training and test errors during training on miniImageNet in Figure~\ref{loss}. It can be observed that the curves of training error are similar, while the curves of test errors are different. While contrastive regularization helps the model generalize better, our auto-view module further improves it. This strongly supports our motivation that encoding fine-grained semantic contents can help metric-based meta-learning  generalize better to novel classes in FSL.

\section{Conclusion}
In this paper, we propose view-learnable contrastive learning to improve few-shot image recognition. In particular, we  design a learning-to-learn algorithm to adaptively learn the views. We carry out two paths of tasks, one is label-guided metric-based meta-learning, another is instance-level classification for exploring fine-grained semantic structure of feature space.  Extensive experiments on benchmarks demonstrate that our method effectively boosts performance of metric-based meta-learning.


 \subsubsection*{Acknowledgments}
This work was partially 
supported by the National Key Research and Development Program of China (No. 2018AAA0100204) and a fundamental Program of  Shenzhen Science and Technology Innovation Commission (No. ZX20210035).

\bibliographystyle{IEEEtran}

\appendix

\section{Learning Procedure}
The pseudo code of our learning procedure is shown in Algorithm\ref{alg}, 

\begin{algorithm}[H]
\caption{View-Learnable Contrastive Learning(VLCL) for metric-based meta-learning}
\label{alg}
\begin{algorithmic}[1]
    \STATE{\textbf{Require}: Base dataset $\mathbf{D}_B$, learning rate $\alpha$ and $\eta$, weight hyperparameter $\beta$, momentum coefficient $\epsilon$, and maximum iteration number $t_{\mathrm{max}}$}
    \STATE{Random initialization for $\theta^0, \omega^0, \gamma^0$}
    \FOR{$t=0$ to $t_{\mathrm{max}}$}
        \STATE{/*  Sample tasks  */\\
        Randomly sample N classes from $\mathbf{D}_B$.\\
        Randomly sample K images from each class in $\mathbf{D}_B$ to form $\mathbf{D}_S^t$\\
        Randomly sample other M images from the same N classes in $\mathbf{D}_B$ to form $\mathbf{D}_Q^t$\\}
        \STATE{ /* First forward pass */\\
        Using $\theta^t$ and $\omega^t$ to compute $\mathcal{L}^{\mathrm{meta}}$ and $\mathcal{L}^{\mathrm{con}}$ through Eq.~(1) and Eq.~(2)}\\
        \STATE{/* Optimize main encoder $F_\theta(\cdot)$, project head $g_\theta(\cdot)$ and momentum encoder $F_\omega(\cdot)$ */\\
        Update $(\theta^t,\omega^t)$ to $(\theta^{t+1},\omega^{t+1})$ through Eq.~(3) and Eq.(4), and retain computational graph.}
        \STATE{ /* Second forward pass */\\
        Using $\theta^{t+1}$ to compute $\mathcal{L}^{\mathrm{meta}}$ through Eq.~(1)}\\
        \STATE{/* Optimize spatial transformation module $G_{\gamma_1}(\cdot)$ and $G_{\gamma_2}(\cdot)$ */\\
        Update $(\gamma^t_1,\gamma^t_2)$ to $(\gamma^{t+1}_1,\gamma^{t+1}_2)$ through Eq.~(5) }\\
    \ENDFOR
\end{algorithmic}
\end{algorithm}

\section{Spatial Transformer Networks}
\label{STN}
In spatial transformer networks~\cite{DBLP:conf/nips/JaderbergSZK15}, the input source image $\mathbf{x}^s$ is first fed into a localisation net ${G}_{\gamma}(\cdot)$ and outputs six affine transformation parameters. This parameters form a 2$\times$3 matrix which defines a affine transformation mapping each pixel coordinates $(u^t_i,v^t_i)$ in the output $\mathbf{x}^t$ to a source coordinates $(u^s_i,v^s_i)$ in the input.   In our setting, we contrain the matrix to be diagonal so as to avoid skewing which could possibly change the semantics of images:
$\{\lambda_{i,j}\}_{i,j\in\{1,2,3\}}$.
\[
\begin{pmatrix}
u^s_i \\ v^s_i
\end{pmatrix}= \tau_\lambda(u^s_t,v^s_t) = 
\begin{bmatrix}
\lambda_{11} & 0 & \lambda_{13} \\
0 & \lambda_{22} & \lambda_{23}
\end{bmatrix}
\begin{pmatrix}
u^t_i \\ v^t_i \\ 1
\end{pmatrix}
\]
Finally, the values of each pixels in $\mathbf{x}^t$ is determined by bilinear interpolation at their corresponding coordinates in the source images, called differentiable image sampling. 

\section{Momentum Contrast}

Different from the standard framework in SimCLR~\cite{DBLP:journals/corr/abs-2002-05709} ,momentum contrast~\cite{DBLP:conf/cvpr/He0WXG20} framework introduces a queue q preserving negative samples and a momentum encoder  $F_\omega(\cdot)$, to alleviate the problem of need for very large batch size for contrastive learning.  In each iteration, immediate preceding features in the queue encoded by $F_\omega(\cdot)$ could be reused as negative samples to compute the contrastive loss. At the end of each iteration, features of current mini-batch is enqueued to the queue, and earliest features in the queue are removed. The update of the encoder $F_\omega(\cdot)$ is intractable by back-propagation. To maintain consistency, MoCo updates $\omega$ as a moving average of the main encoder's parameter $\theta$, as shown in eq. (4).

\section{Results for few-shot fine-grained classification}
Fine-grained categories are distinguished by subtle and local semantic differences, which makes few-shot fine-grained classification more difficult. We experimentally show that such difficulty can be largely addressed by our method. Table 1 presents 5-way mean accuracy on three datasets with fine-grained categories: Cars~\cite{DBLP:conf/iccvw/Krause0DF13}, Places~\cite{DBLP:journals/pami/ZhouLKO018} and Plantae~\cite{DBLP:conf/cvpr/HornASCSSAPB18}. It can be observed that our method improved performance of Prototypical Network by a large margin under both 5-shot and 1-shot settings. For instance, our method obtains 14.83\% and 11.21\% gains under 1-shot and 5-shot settings on Cars, respectively. This verifies that metric-based meta-learning benefits from better fine-grained semantic structure learnt by our method.
\begin{table}[t]
\footnotesize
\caption{Results for 5-way few-shot classification on three fine-grained datasets: Cars, Places and Plantae. Average accuracies on the meta-test set are reported.}
    \label{fine-grain}
    \begin{center}
    \begin{tabular}{c||c|c|c|c|c|c}
    \multirow{3}{*}{Model} &  \multicolumn{2}{c|}{Cars} & \multicolumn{2}{c|}{Places} & \multicolumn{2}{c}{Plantea} \\
    
    & 1-shot & 5-shot & 1-shot & 5-shot & 1-shot & 5-shot \\ 
    \hline
    
    PN & 62.11 & 75.83 & 62.19 & 76.67 & 53.59 & 67.98 \\
    PN+VLCL & \textbf{76.93} & \textbf{87.04} & \textbf{64.50} & \textbf{78.96} & \textbf{59.80} & \textbf{75.37}  \\ \hline

    \end{tabular}
    \end{center}
\end{table}

\section{Effect of regularization hyperparameter}
We perform ablation study w.r.t. the regularization hyperparameter $\beta$ which controls the magnitude of contrastive loss. Table 2 shows the accuracies of 5-way few-shot learning on miniImageNet. We can observe that when $\beta$ is small, the contrastive loss cannot thoroughly explore the semantics inside data, thus cannot boost the performance much. When the value of $\beta$ is s too large, the accuracy also decreases. This implies that supervised information is somewhat ignored, which is  still important.

\begin{table}[t]
\caption{Effects of the $\beta$ value in VLCL on model performances.}
    \label{beta}
    \begin{center}
    \begin{tabular}{c||c|c}
    & \multicolumn{2}{c}{\textbf{5-way Acc.}}\\
    & 1-shot & 5-shot \\\hline
    $\beta$ = 0.5 & 58.84 $\pm$ 0.39\% & 74.21 $\pm$ 0.55\%\\
    $\beta$ = 1.0 & 59.45 $\pm$ 0.36\%& 75.64 $\pm$ 0.71\%\\
    $\beta$ = 2.0 & 61.75 $\pm$ 0.43\% & 77.19 $\pm$ 0.51\%\\
    $\beta$ = 5.0 & 58.73 $\pm$ 0.50\% & 74.37 $\pm$ 0.77\% \\\hline
    
    \end{tabular}
    \end{center}
\end{table}
\end{document}